\pdfoutput=1

\documentclass[11pt]{article}

\usepackage[final]{acl}
\usepackage{times}
\usepackage{latexsym}
\usepackage{tikz}
\usepackage[T1]{fontenc}

\usepackage[utf8]{inputenc}

\usepackage{microtype}

\usepackage{inconsolata}

\usepackage{graphicx}
\usepackage{multirow}
\usepackage{float}
\title{KnowLab\_AIMed at MEDIQA-CORR 2024: Chain-of-Though (CoT) prompting strategies for medical error detection and correction}

\author{Zhaolong Wu\textsuperscript{1}\thanks{These authors contributed equally to this work.}, 
        Abul Hasan\textsuperscript{2}\footnotemark[1],\\ 
        \textbf{Jinge Wu\textsuperscript{2}}, 
        \textbf{Yunsoo Kim\textsuperscript{2}}, 
        \textbf{Jason P.Y. Cheung}\textsuperscript{1}\thanks{Corresponding authors.}, 
        \textbf{Teng Zhang}\textsuperscript{1}\footnotemark[2], 
        \textbf{Honghan Wu}\textsuperscript{2}\footnotemark[2]\\
\textsuperscript{1}Department of Orthopaedics and Traumatology, University of Hong Kong\\
\textsuperscript{2}Institute of Health Informatics, University College London\\
\texttt{\{wuzl01\}@connect.hku.hk}, \\
\texttt{\{cheungjp, tgzhang\}@hku.hk}, \\
\texttt{\{a.kalam, jinge.wu.20, yunsoo.kim.23, honghan.wu\}@ucl.ac.uk}\\ 
}

\tikzstyle{mybox} = [draw=red, fill=blue!20, very thick,
    rectangle, rounded corners, inner sep=10pt, inner ysep=20pt]
\begin{document}
\maketitle
\begin{abstract}
This paper describes our submission to the MEDIQA-CORR 2024 shared task for automatically detecting and correcting medical errors in clinical notes. We report results for three methods of few-shot In-Context Learning (ICL) augmented with Chain-of-Thought (CoT) and reason prompts using a large language model (LLM). In the first method, we manually analyse a subset of train and validation dataset to infer three CoT prompts by examining error types in the clinical notes. In the second method, we utilise the training dataset to prompt the LLM to deduce reasons about their correctness or incorrectness. The constructed CoTs and reasons are then augmented with ICL examples to solve the tasks of error detection, span identification, and error correction. Finally, we combine the two methods using a rule-based ensemble method. Across the three sub-tasks, our ensemble method achieves a ranking of 3rd for both sub-task 1 and 2, while securing 7th place in sub-task 3 among all submissions.     
\end{abstract}

\section{Introduction}
The rise of Large Language Models (LLMs) such as  GPT4 \cite{achiam2023gpt}, Med-PaLM \cite{singhal2023large}, and LLaMA \cite{touvron2023llama, touvron2023llama2} have inspired investigations into their potential use in automatically analysing Electronic Health Records (EHRs). However, the usefulness of LLMs in clinical settings remains challenging due to the fact that these models are trained on large-scale corpora which may contain inaccuracies, common mistakes, and misinformation \cite{thirunavukarasu2023large, ji2023survey}. To motivate research on the problem of identifying and correcting common sense medical errors in clinical notes using LLMs, the MEDIQA-CORR (Medical Error Detection \& Correction) shared tasks are proposed. Herein, we describe our submissions to the shared tasks presenting two methodologies and an ensemble approach using GPT4, all utilising In-Context Learning (ICL) \cite{icl_Brown}  in conjunction with Chain-of-thought (CoT) \cite{reasoning, wang2022self} and reason prompts. The ensemble method achieves accuracies of 69.40\% and 61.94\% for sub-task 1 and sub-task 2, respectively, while obtaining a BLUERT score of 0.6541 for sub-task 3.      
\section{Shared Tasks and Dataset}
\subsection{Shared Tasks}
The MEDIQA-CORR 2024\cite{mediqa-corr-task} proposes three sub-tasks:
\begin{enumerate}
    \item \textbf{Binary Classification (sub-task 1)}: To detect whether a clinical note contains a medical error.
    \item \textbf{Span Identification (sub-task 2)}: To identify the text span (i.e. Error Sentence ID) associated with the error, if a medical error exists in the clinical note. 
    \item \textbf{Natural Language Generation (sub-task 3)}:
    To generate a corrected text span, if a medical error exists in the clinical note.
\end{enumerate}
\subsection{Dataset}
The training dataset is derived from a single source called as MS Training Set, where as the validation and test datasets are derived from two different sources termed as MS and UW Validation/Test set \cite{mediqa-corr-dataset}. The MS Training Set is comprised of 2,189 clinical notes. The MS Validation Set includes 574 clinical notes, while the UW Validation Set includes 160 clinical notes. The Test dataset has in total 926 clinical notes derived from two sources. 

\section{Methods}
  \begin{figure*}[!h]
  \centering
  \includegraphics[scale=0.65]{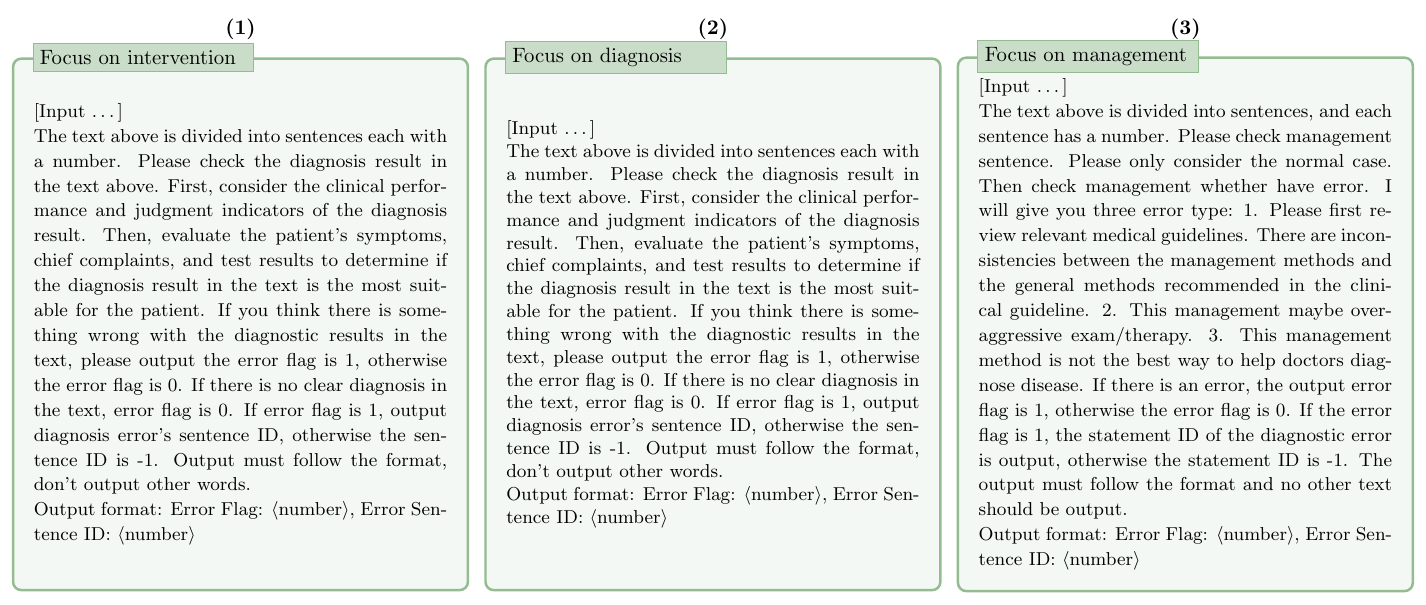}
  \caption{Three types of Chain-of-Thought (CoT) prompts utilised in the ICL-RAG-CoT method: (1), (2), and (3) direct the GPT4 model to focus on intervention, diagnostic, and management errors, respectively.}
  \label{fig:01}
\end{figure*}
\subsection{ICL-RAG- augmented with CoT prompting(ICL-RAG-CoT)}
The Chain-of-Thought (CoT) prompting method, which includes a sequence of reasoning steps, has demonstrated enhancements in the problem-solving capabilities of LLMs over standard prompting techniques, particularly in solving mathematical tasks \cite{reasoning, kojima2022large, yao2024tree}. Recent studies, such as the one conducted by \cite{kim2023cot}, have introduced datasets that incorporate CoT instructions aimed at addressing various Natural Language Processing (NLP) tasks. These tasks include question answering and natural language inference and have been tailored for smaller-scale language models like Flan-T5 \cite{longpre2023flan}. Motivated by these developments, we conduct a manual analysis of a subset derived from both the MS Training set and UW Validation set to investigate the prevalent error types within clinical notes. Our examination reveals three broad categories of errors evident in the clinical notes; they are : (1) Diagnosis, (2) Intervention, and (3) Management. Using these categories we construct three separate prompts, shown in Figure ~\ref{fig:01}, that are augmented with ICL examples.\\
To address the three sub-tasks, our initial approach, referred to as ICL-RAG augmented with CoT prompting (ICL-RAG-CoT), adopts a two-stage prompting methodology with GPT4. For the binary classification and span identification tasks (i.e. sub-task 1 and sub-task 2), we guide GPT4 systematically through a sequence of prompts, each tailored to detect and identify medical errors. The first prompt in the sequence is a standard prompting which tasks the model to detect errors in a clinical note, supplemented with in-context examples. If no medical error is detected, we proceed to prompt GPT4 iteratively by augmenting our CoTs in Figure ~\ref{fig:01} with ICL examples until an error is identified. Once all CoTs are exhausted, the clinical note is considered error-free.
In the second stage, for the NLG task, we prompt GPT4 independently by specifying the predicted incorrect sentence number (i.e., Sentence ID) obtained from the first stage. A prompt template is provided in Appendix ~\ref{sec:appendix_a}; see Figure ~\ref{fig:04}. In order to generate In-context examples for prompting LLMs, our methodology incorporates the Retrieval-Augmented Generation (RAG) approach, as proposed by \citet{rag_lewis, rag_jin}. Utilising the e5-large-unsupervised model \citep{e5}, we transform the MS-Training dataset into a vectorized database. This process involves applying cosine similarity to find the $k$-most similar training instances for each validation and test input. In our experiments we select $k$=4. 
 \begin{figure*}[!h]
  \centering
  \includegraphics[scale=0.9]{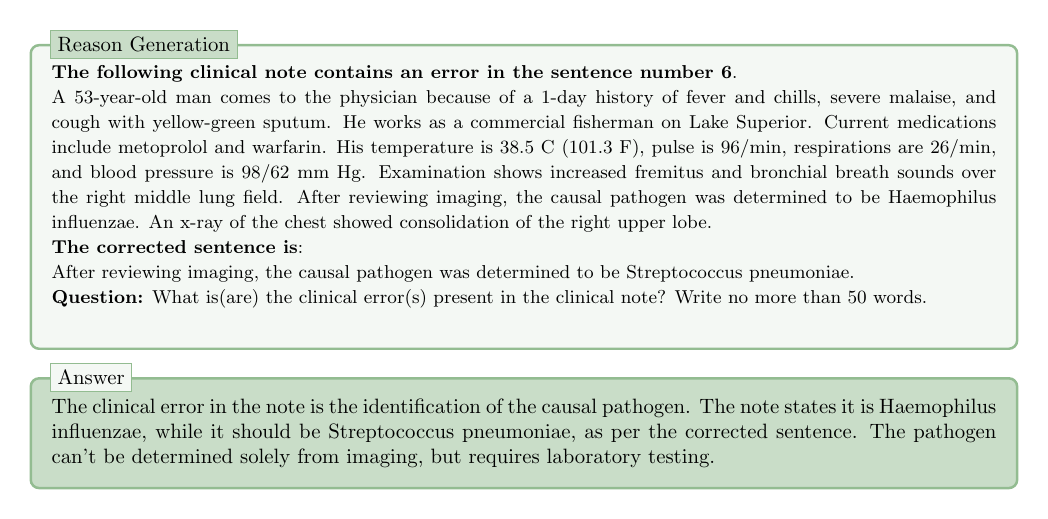}
  \caption{Reason generation template utlised in the ICL-RAG-Reason method}
  \label{fig:02}
\end{figure*}
\subsection{ICL-RAG- augmented with reason (ICL-RAG-Reason)}
 In our second method, referred to as ICL-augmented with reason (ICL-RAG-Reason), we aim to address three sub-tasks simultaneously using a single prompt containing ICL examples and their corresponding reasons for correctness or incorrectness. However, this method requires to prompt the LLM to pre-process the training data separately. Consequently, the ICL-RAG-Reason method begins by prompting GPT4 to generate a brief reason for the correctness or incorrectness of a clinical note from the MS Training set; see Figure ~\ref{fig:02} for an example. If a note contains an error, we prompt the LLM by concatenating it with the corrected sentence to explain why the clinical note is deemed incorrect. In the case of a correct training example, we prompt the GPT4 to provide us with the clinical characteristics that validate the note's correctness. Thus, we automatically construct reasoning instructions for each MS Training notes. We employ a similar RAG method to ICL-RAG-CoT; however, we utilize OpenAI embeddings \footnote{\url{https://platform.openai.com/docs/guides/embeddings}} to embed all clinical notes across the three datasets. For every input validation and test note, we sample 4 (4-shot) training notes from a pool of its semantically most similar $k$ notes, comprising two correct and two incorrect notes. We augment selected training notes with their {\em Reasons} for being correct or incorrect and create the final prompt; ; see Figure ~\ref{fig:05} in Appendix ~\ref{sec:appendix_a} for an example of prompt template. The ICL-RAG-Reason method samples ICL examples three times to ensure that the model is shown different reasoning paths. This sampling strategy provides us with three different solutions which is resolve by majority voting to ensure consistency and then take the corrected sentence by randomly selecting one from two correct answers.
 \begin{table*}[!h]
\centering
\caption{Main results. Here Acc, AG, R1, and AGC denote Accuracy, Aggregate, ROUGE-1, and AggregateC scores, respectively. }
\label{tab:tab01}
\begin{tabular}{|l|l|l|l|l|l|l|l|} \hline
&Sub-task 1&Sub-task 2&\multicolumn{5}{c|}{Sub-task 3} \\ \cline{2-8} 
Method &Acc&Acc&AG& R1 & BERT & BLEURT & 
AGC \\ \hline
\textbf{MS Validation}&&&&&&&\\ \cline{1-1} 
ICL-RAG-CoT&\textbf{0.6620}&\textbf{0.6236}&\textbf{0.6350}&\textbf{0.6028}&\textbf{0.6658}&\textbf{0.6363}&\textbf{0.5067}\\ \hline
ICL-RAG-Reason&0.6010& 0.5644&0.6165&0.5739& 0.6577&  0.6178& 0.4298 \\ \hline 
Ensemble &  \textbf{0.6620}&\textbf{0.6236}& 0.6184 & 0.5777 & 0.6560 &0.6215&0.5048 \\ \hline
\textbf{UW Validation}&&&&&&&\\ \cline{1-1} 
ICL-RAG-CoT&\textbf{0.7437}&\textbf{0.6500}& 0.6525&0.6701&0.6519& 0.6355&0.6091\\ \hline
ICL-RAG-Reason&0.6875& 0.5625&0.6340&0.6180& 0.6343&  0.6499&0.5350 \\ \hline 
Ensemble & \textbf{0.7437}&\textbf{0.6500}&\textbf{0.6740} &\textbf{0.6762} & \textbf{0.6729} &\textbf{0.6728}&\textbf{0.6174} \\ \hline
\textbf{Test}&&&&&&&\\ \cline{1-1} 
ICL-RAG-CoT&\textbf{0.6940}&\textbf{0.6194}&0.6255&0.6130& 0.6399& 0.6235&0.5346 \\ \hline
ICL-RAG-Reason&0.6540& 0.5837&0.6509&0.6343& 0.6703& 0.6482&0.5119 \\ \hline 
Ensemble & \textbf{0.6940}&\textbf{0.6194}&\textbf{0.6581} & \textbf{0.6434} & \textbf{0.6767} & \textbf{0.6541} & \textbf{0.5730} \\ \hline 
\end{tabular}
\end{table*}

\subsection{Ensemble}
We integrate the ICL-RAG-CoT and ICL-RAG-Reason methods using a rule-based approach, henceforth termed as the Ensemble method. This approach initially considers predictions generated by the ICL-RAG-CoT method for sub-task 1 and sub-task 2 as correct, while predictions for sub-task 3 from ICL-RAG-Reason are also deemed correct. It then resolves conflicts by identifying clinical notes from the MS and UW Validation and Test sets that are predicted as incorrect by both methods but have differing Error Sentence IDs. Finally, the Ensemble method prompts GPT4 (see see Figure ~\ref{fig:06} in Appendix ~\ref{sec:appendix_a} for an example), providing it with ICL examples, each containing an error, to generate a corrected sentence by specifying the Eorror Sentence ID predicted by the ICL-RAG-CoT. 
\subsection{Evaluation}
We evaluate the performances of our methods with the official evaluation scripts on MS and UW Validation Set \footnote{\url{https://github.com/abachaa/MEDIQA-CORR-2024}}. Sub-task 1 and 2 are evaluated by using Accuracy. The Natural Language Generation task (i.e. sub-task 3) is evaluated with 
with ROUGE \cite{lin-2004-rouge}, BERTScore \cite{zhang2019bertscore}, and BLEURT \cite{sellam-etal-2020-bleurt}. We report performances as the arithmetic mean of ROUGE-1 F1, BERTScore, BLEURT-20. Furthermore, Aggregate scores and AggregateComposite scores, the overall measures across the mentioned metrics, are provided.
\section{Results}
We attain accuracies of 66.20\%, 74.37\%, and 69.40\% on the MS Validation, UW Validation, and Test datasets, respectively, for the binary classification task of error detection (i.e. sub-task 1) using the ICL-RAG-CoT method; see Table ~\ref{tab:tab01}. For the span identification task, i.e. sub-task 2, the same method achieves accuracies of 62.36\%, 65.00\%, and 61.94\%, respectively. It is noteworthy that the Ensemble method achieves similar accuracies. In the sub-task 3, which involves Natural Language Generation (NLG), the ICL-RAG-CoT method performs less effectively compared to the ICL-RAG-Reason method. It reaches a BLEURT score of 0.6363 on the MS Validation Set. However, our Ensemble approach surpasses the other two methods, achieving BLEURT scores of 0.6729 and 0.6541 for the UW Validation and Test sets, respectively. We observe similar perfomances across other NLG metrics; see Table ~\ref{tab:tab01}. This is because the reasoning generation method. i.e. ICL-RAG-Reason achieves better performances than the ICL-RAG-CoT method particularly in the NLG task. 
\begin{figure}[!h]
  \centering
  \includegraphics[scale=0.65]{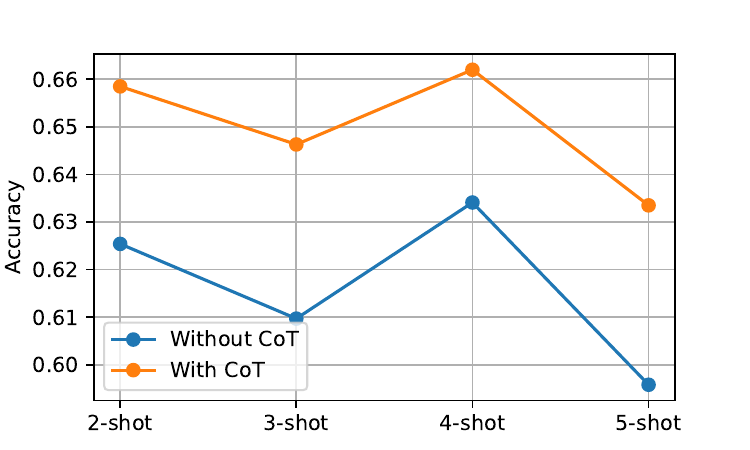}
  \caption{Comparison of few-shot examples with or without CoT using ICL-RAG-CoT method on the Binary Classification Task (i.e. sub-task 1) on the MS Validation Set}
  \label{fig:03}
\end{figure}
\section{Discussion}
Our CoT prompting strategy works well in conjunction with the RAG system. As depicted in Figure ~\ref{fig:03}, across various few-shot settings (e.g., 2, 3, 4, and 5-shot settings), the ICL-RAG-CoT method consistently outperforms scenarios where CoT is not employed alongside RAG in the binary classification task. We observe that both the 3-shot and 5-shot settings yield lower performance compared to the 2-shot and 4-shot settings. This disparity suggests that class imbalance in few-shot settings could potentially deteriorate performance. This motivates our selection of 4-shot setting consistently across all our experiments. One of the limitations of our study is that we do not rigorously evaluate the NLG Task, i.e. sub-task 3. Consequently, our overall ranking falls towards the lower end of the top 10 (ranked 7 over-all). While our Ensemble prompting strategy demonstrates a good  performance by leveraging reasoning gathered independently from GPT4, there remains scope for improvement. For instance, further enhancement could be achieved by evaluating the generation of LLMs against clinical and/or biomedical knowledge bases to verify their output.
\section{Conclusion}
 We present our submission to the MEDIQA-CORR shared task for medical error detection and correction. Our study evaluates the effectiveness of the GPT4 model through various prompting strategies employing CoT prompting and Reasoning methods. Specifically, our CoT prompting strategies achieve high accuracies in error detection and identification tasks. Additionally, our Ensemble method, which combines outputs from both methods, demonstrates a better performance on the NLG task than the CoT prompting alone. In the future, we aim to explore our approach for other downstream tasks in the clinical domain using open-source LLMs.

\section{Ethical Statement}
Our research employs large language model (LLM) to improve the accuracy of medical records. However, before deploying and utilising the methods proposed with LLM, it is necessary to adhere to ethical and moral principles. The storage and use of patient data must strictly comply with data protection and privacy laws, such as Health Insurance Portability and Accountability Act (HIPAA) and General Data Protection Regulation (GDPR), to ensure that data access is strictly controlled and process transparency is maintained.
\section{Acknowledgement}
Abul Hasan and Honghan Wu are funded by the National Institute for Health Research (NIHR) Artificial Intelligence and Multimorbidity: Clustering in Individuals, Space and Clinical Context (AIM-CISC) grant NIHR202639. The views expressed are those of the authors and not necessarily those of the NIHR or the Department of Health and Social Care.
\bibliography{main}
\newpage
\appendix
\section{Prompt Templates}
\label{sec:appendix_a}
\begin{figure}[!h]
  \centering
  \includegraphics[scale=0.6]{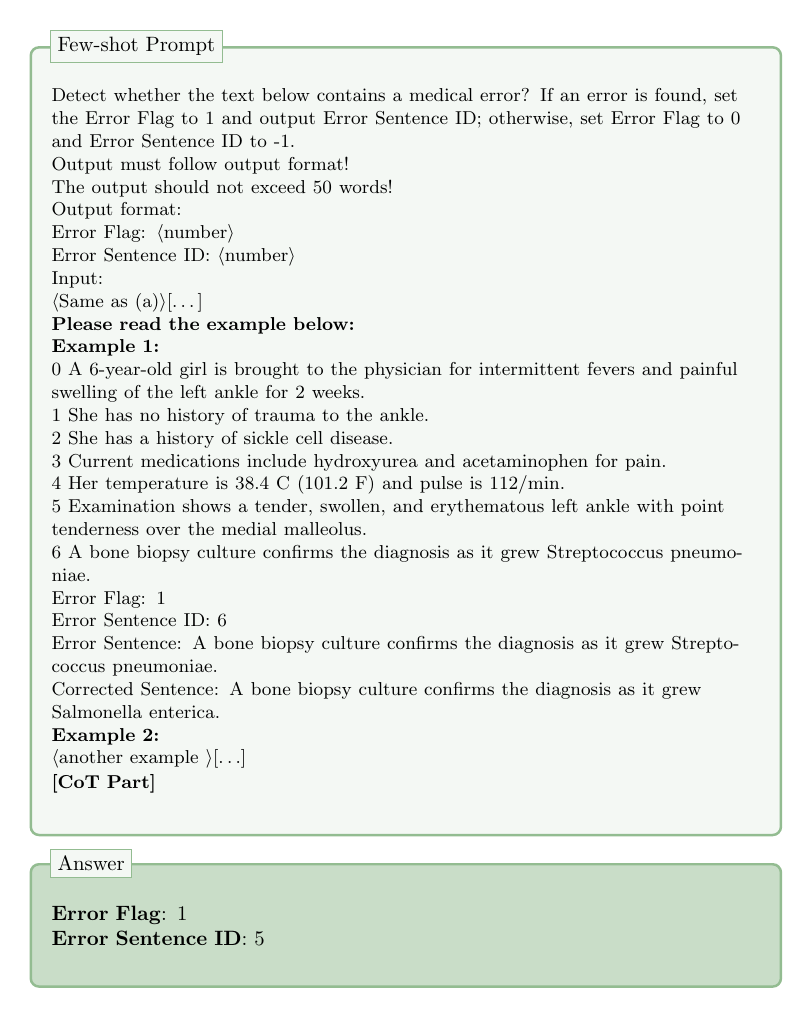}
  \caption{A template used in ICL-RAG-CoT for the few-shot prompting to solve sub-task 1 and 2.}
  \label{fig:04}
\end{figure}
\begin{figure}[!ht]
  \centering
  \includegraphics[scale=0.6]{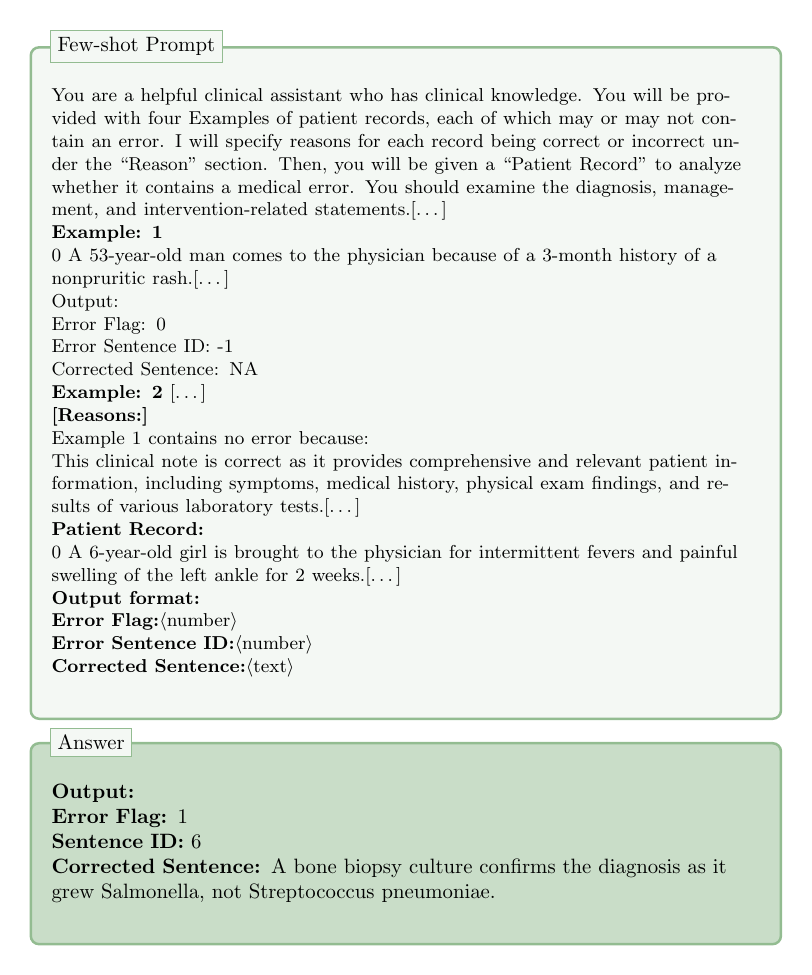}
  \caption{A template used in ICL-RAG-Reason for the few-shot prompting to solve all sub-tasks simultaneously.}
  \label{fig:05}
\end{figure}
\begin{figure}[!h]
  \centering
  \includegraphics[scale=0.6]{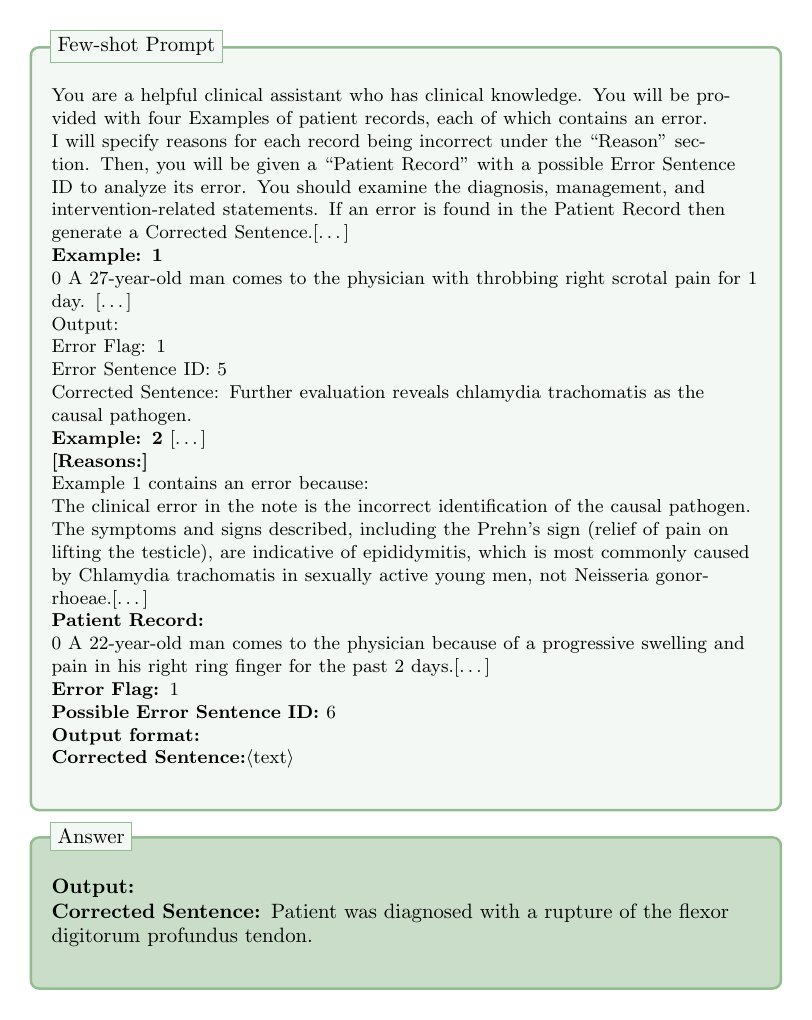}
  \caption{A template used in Ensemble method for the few-shot prompting to solve the sub-task 3.}
  \label{fig:06}
\end{figure}
\end{document}